# An Empirical Testing of Autonomous Vehicle Simulator System for Urban Driving

**John Seymour**
Department of Computing Technologies
Swinburne University of Technology
Hawthorn, VIC 3122, Australia
john.seymour787@gmail.com

**Dac-Thanh-Chuong Ho**
Department of Computing Technologies
Swinburne University of Technology
Hawthorn, VIC 3122, Australia
hodacthanhchuong@gmail.com

**Quang-Hung Luu** *
Department of Computing Technologies
Swinburne University of Technology
Hawthorn, VIC 3122, Australia
hluu@swin.edu.au
luuquanghung@gmail.com

*Abstract*—Safety is one of the main challenges that prohibit autonomous vehicles (AV), requiring them to be well tested ahead of being allowed on the road. In comparison with road tests, simulators allow us to validate the AV conveniently and affordably. However, it remains unclear how to best use the AV-based simulator system for testing effectively. Our paper presents an empirical testing of AV simulator system that combines the SVL simulator and the Apollo platform. We propose 576 test cases which are inspired by four naturalistic driving situations with pedestrians and surrounding cars. We found that the SVL can imitate realistic safe and collision situations; and at the same time, Apollo can drive the car quite safely. On the other hand, we noted that the system failed to detect pedestrians or vehicles on the road in three out of four classes, accounting for 10.0% total number of scenarios tested. We further applied metamorphic testing to identify inconsistencies in the system with additional 486 test cases. We then discussed some insights into the scenarios that may cause hazardous situations in real life. In summary, this paper provides a new empirical evidence to strengthen the assertion that the simulator-based system can be an indispensable tool for a comprehensive testing of the AV.

## I. Introduction

Autonomous vehicles (AV) are promised to provide convenient, economical, and safe trips. It is no doubt that the AV is the future of our transportation. One major concern that prohibits the AV to be on the road is its safety. An AV has killed a pedestrian in Tempe, Arizona in 2016 when she was walking and then hit by an AV [1]. Several fatal incidents caused by the malfunction of AV have been reported since then. For that reason, assuring the quality of AV is of great importance.

Having said that, the road testing of AV to assure its quality is often very costly. Besides, the testing may not be reproducible due to the difference in physical environments as well as parties and entities involved in the test. Road tests also involve regular traffic and, in many cases, are unsafe for the people involved. The use of simulators becomes increasingly important for testers as it can provide a cheap and reproducible way to examine the quality of AV [2]. Recent studies indicated that the simulator-generated scenes are as accurate as real-life datasets in deep learning models that drive the AV [3]. Hence, simulators can be harnessed to support the testing of AVs effectively and efficiently.

Both road tests and simulator-based tests face the same challenges associated with. As highlighted by the 2021 IEEE Autonomous Driving AI Test Challenge, they are: the absence of well-defined validation standards and criteria for testing AVs; and the lack of automation tools to generate test scenarios cost-effectively. To partially address this challenge, our paper presents new empirical evidence that it is feasible to (i) derive a good scenario diversity, and at the same time (ii) detect AV problems in the simulation environment. In addition, we can take advantage of the experiments to gain insights into hazardous situations in reality.

In this paper, we prepare four different classes of scenarios for testing the AV simulator system that combines SVL and Apollo. Each class of scenario is inspired by commonly observed naturalistic driving situations with pedestrians and surrounding cars. We specify that the AV system is safe to drive the ego car out of hazardous scenarios. In testing, we adopt the Equivalent Partitioning (EP) and Metamorphic Testing (MT) techniques [4]. MT has been applied successfully in detecting problems in some offline AV systems [5], which make independent decisions based on static photo inputs. In contrast, online testing is the close-loop mode in which there are interactions with the application environment [3]. The adoption of simulator may allow us to examine whether MT is still effective in detecting issues in the online mode. We focus on urban driving where there are more people as well as more interactions between pedestrians and vehicles.

Our empirical investigation is aimed to partially address the following research questions: RQ1: Are our test cases effective in mimicking hazardous situations and can the system capture them? RQ2: What are critical factors that distinguish a scenario to be safe or unsafe? RQ3: Does the environmental condition have any impact on the safety of the AV? RQ4: Is the use of simulator-based systems helpful in understanding and guaranteeing the safety in reality? We organise the paper as follows. We first describe our methodology and selection of test scenarios in Section II. Experiments including our system under test and the generation of test cases are shown in Section III. We present the main results in Section IV. We further discuss their insights in Section V. The concluding remark in Section VI will then summarise our main findings.

## II. Methodology

### A. Testing methodology

In this study, we adopt two black-box testing strategies. The advantage of black-box testing is the ability to reuse the test cases for different simulators and platforms. We focus on

* *corresponding author.*
This work is supported in part by the grant ARC DP210102447. The paper has been awarded the Third Prize in the 2021 IEEE Autonomous Driving AI Test Challenge.

XXX-X-XXXX-XXXX-X/XX/$XX.00 ©20XX IEEE

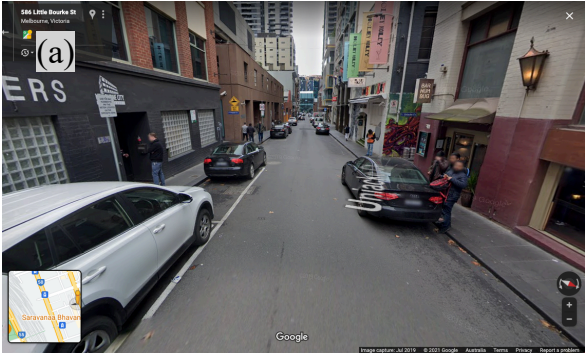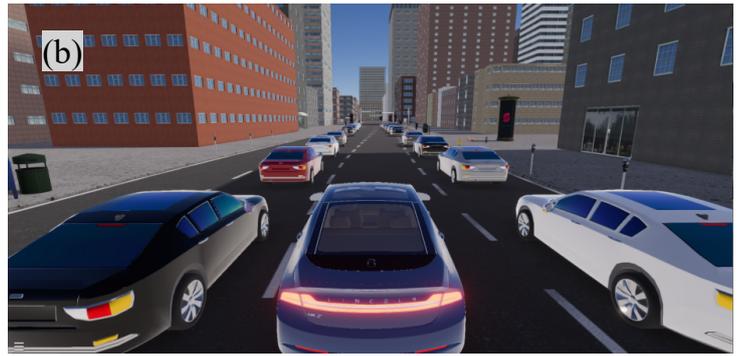

Fig. 1. **(a)** Scenario Class inspired by a naturalistic street scene with cars and pedestrians in Little Bourke Street (Melbourne) observed from Google Streetview [copyright by Google], and **(b)** a setting of NPCs and pedestrians in our experiment with SVL simulator.

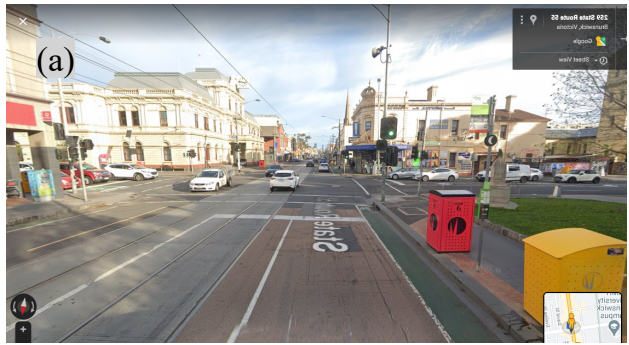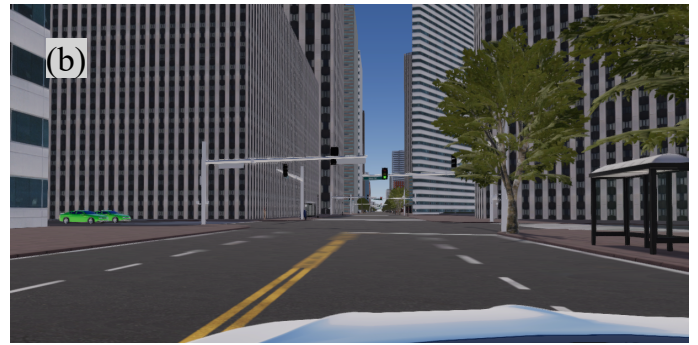

Fig. 2. **(a)** Scenario Class B inspired by a naturalistic street scene (flipped) with cars and pedestrians at Sydney Road (Melbourne) observed from Google Streetview [copyright by Google] and **(b)** a setting of NPC and pedestrian in our experiment with SVL simulator.

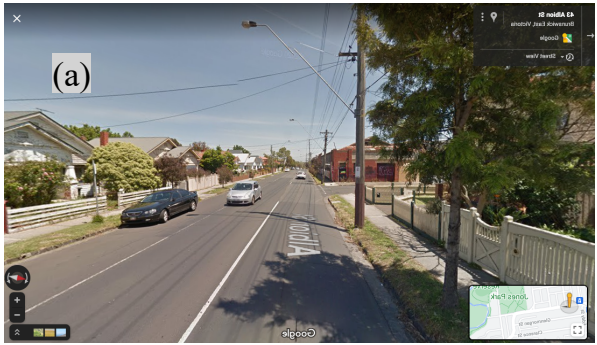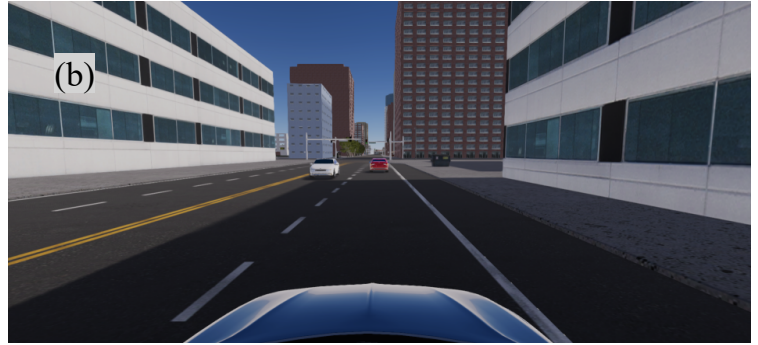

Fig. 3. **(a)** Scenario Class C inspired by a naturalistic street scene (flipped) where the ego car is blocked by other car at Albion Street (Melbourne) as observed from Google Street view [copyright by Google], and **(b)** a setting of NPCs in our experiment with SVL simulator.

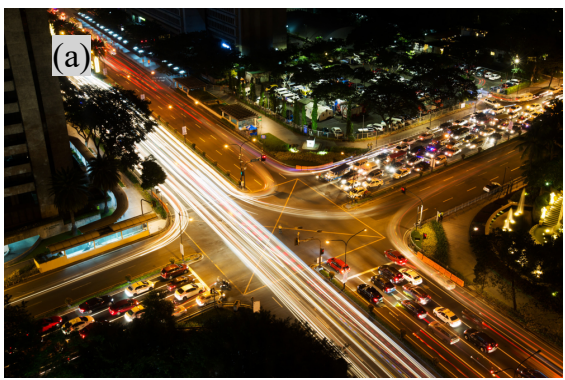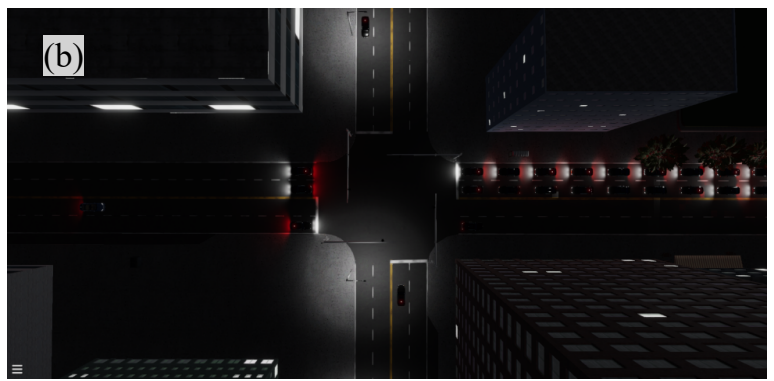

Fig. 4. **(a)** Scenario Class D inspired by a naturalistic night street scene at an intersection with cars [copyright by SeongJoon Cho / Bloomberg via NBC News/Getty Images], and **(b)** setting of NPCs in our experiment with SVL simulator.

the validation, that is, determine whether the system can capture the real-life situations – instead of working on the verification. In line with this, in the first strategy, we set up the test cases based on the EP technique with a focus on valid inputs only. We assume the EP domain is confined by independent dimensions for each class of scenarios. Each parameter is designed to have 2-4 ranges (to have, say, small, medium, and large values). Therefore, an output can be constructed by a combination of composite parameters. For example, if we have 2 parameters, each has three values (represented by its corresponding three ranges), then the output consists of 9 (=3x3) EP domains. For each EP, we select a certain value in the range to construct a test case. In the specification, the AV system is required to drive the car away from hitting the pedestrians or non-pedestrian characters (NPC). Otherwise, the test is considered a failure when collision occurs.

In the second strategy, we adopt the metamorphic technique. This strategy allows us to further detect the inconsistencies of the system. Given a source input $I_s$, which is a chosen scenario setup with the NPC. Assume we can derive a follow-up input $I_f$, which is the same as the source input, except that the weather, road, or timing condition is different. The system will have two different returned states, referred to as the source output $O_s$, and the follow-up output $O_f$, respectively. The output could be a decision, a series of decisions that affect the ego car's speed, steering angle, location, state of the ego car, traffic conditions that affect the car, or collisions and safe situations. Our metamorphic relations (MR) are derived from the intuition that the system driving the ego car should make a consistent decision; that is, regardless of the weather condition, the output should be the same. If there is a difference between the two outputs, the system is determined to be faulty. In this preliminary study, we simply adopt the collision or safe situation outcomes. For example, if the source scenario during daytime is safe but the follow-up scenario during night-time gives a collision, the MR is considered violated, and hence the system is faulty. Thanks to the use of a simulator, we can have the similar scenario but with a different setup, which cannot be done with the road testing.

*B. Classes of test scenarios*

In this paper, we have four different classes of scenarios associated with urban driving: pedestrian only (A), pedestrian and moving NPCs (B), NPCs only (C), and NPCs only under a complex environment (D).

*a) Scenario Class A (Pedestrians in Close Quarters):* This set of scenarios sees the Apollo ego navigate through a tight space with stationary cars on either side, possibly simulating a narrow street or a traffic jam, as illustrated in Figure 1 above. Suddenly, two pedestrians appear between some cars and cross in front of the ego. The purpose of this test is to attempt to identify if the ego can successfully detect the danger and stop in time. Additionally, if sensor detection is used, this scenario would also "stress" the perception module by providing multiple entities to track, potentially increasing the chance of missing pedestrians.

*b) Scenario Class B (Pedestrian at Intersection):* This set of scenarios also relates to testing Apollo's ability to detect and respond to pedestrians. In this situation, the ego vehicle is set to make a left turn at an intersection. A pedestrian is then triggered to walk across the intersection parallel to the original direction of the ego, directly crossing its path as it makes the turn. Two stationary cars are also placed at the intersection, although their presence serves little direct purpose, only to somewhat block the view of the pedestrian. This class of scenario is shown in Figure 2. The goal of this test is to again identify if Apollo can respond to pedestrians in time. However, different to the first class of scenarios, an intersection with a turn is involved; and the ego expects to have a complete right to move. Additionally, the ego will have more time to recognise and respond to the moving pedestrian. Various values affecting this test scenario are needed to allow it to be parameterised for automated testing, as further discussed in the results. These values are the speed and trigger distance of the pedestrian, as well as its initial starting distance from the road.

*c) Scenario Class C (Go Around, Oncoming):* This scenario is about testing the ego's ability to detect and respond to oncoming vehicles, as it tries to get around a blocked car in its way (Figure 3). In some ways, this scenario is similar to Deliverable 1's Scenario 5 of the Competition, where the ego needs to respond to oncoming vehicles. However, this time the ego vehicle is the one that wants to encroach on to the other side of the road. This scenario uses the Borregas Avenue map as this map has a road with only 1 lane per direction, unlike the San Francisco map. The ego vehicle is placed on one end of the road, with an oncoming NPC vehicle spawned some distance away and in the oncoming adjacent lane, with a trigger distance set to cause it to approach the ego. A second, stationary, NPC is then placed in the same lane as the ego vehicle, a few meters directly ahead. Apollo is then commanded to drive with a destination set just ahead of the stationary NPC. General weather conditions are parameterised, along with the oncoming NPC vehicle speed, spawn distance, and trigger distance.

*d) Scenario Class D (Camera Tricks at Intersection):* The goal of this set of scenarios is to understand and confirm the reaction of the ego vehicle in the task of navigating a busy intersection (Figure 4). This is simulated using many stationary NPCs in the parallel lane to the ego's starting position. Additionally, 2 NPC vehicles are triggered to cross the intersection perpendicularly when the ego approaches, thus getting in its way. The ego vehicle can be set to make a left turn, right turn, or proceed straight ahead through the intersection, depending on the parameter passed. Additional parameters that can be adjusted include the trigger distance and speed for the left and right NPCs individually, as well as the time of day, weather conditions, and NPC car colour. One of the main motivations for creating this specific scenario was to test how well the ego's perception module can handle different environment and lighting conditions. As such, all NPC vehicles are made black by default.

III. EXPERIMENT

*A. System-under-test*

In our experiments, we set up a system that comprises the SVL simulator (previously LGSVL) and the Apollo platform. From our observation, the SVL provides a near-realistic, flexible and efficient 3D simulation environment developed by LG Electronics America R&D Centre. It allows us to customise a wide range of environmental parameters, including maps, characters (pedestrians, vehicles), weather (fog, rain, wetness, time of the day), traffic setup as well as sensor outputs (camera, LIDAR, localisation, etc.). Our ego car is driven by the Apollo system, an AV platform developed by Baidu. The Apollo system incorporates deep learning models to provide driving capabilities. We adopt a recent version of SVL (2021.1) and the latest version of Apollo (6.0) for our experiments.

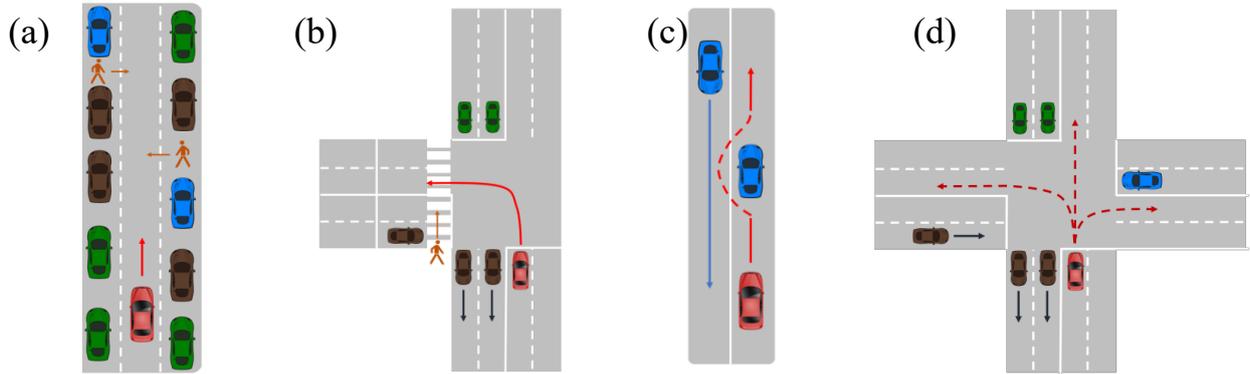

Fig. 5. Illustration of classes of scenarios used: (a) A (Pedestrians in Close Quarters), (b) B (Pedestrians at Intersection), (c) C (Go Around, On Coming NPCs), (d) D (Camera Tricks at Intersection).

All scenarios are created using the San Francisco and Borregas Avenue maps in our setup. To configure them properly, we manually downloaded the map binaries (routing_map.bin, sim_map.bin, base_map.bin) from the Apollo 5.0 LGSVL Fork, before placing them into the appropriate directory.

### B. Test case generations and validation

In this study, we have a total number of four different classes of scenarios, namely A, B, C and D. In each class of scenario, we have a range of parameters. For a unique set of combined parameters, we have a test case. Overall, we have a total number of 576 test cases, each being assigned with a unique scenario (Table II). In addition, we have generated 486 test cases while applying the metamorphic testing, bringing the total number of experimental runs to 1062.

For each scenario, we have coded a Python script and its associated Bash script to do automated testing. We implemented the code in a Bash script to make sure that each test case is not affected by others, the memory burden is reduced, and the overall efficiency in increased. All reproducible codes and results are provided in the GitHub submission.

In the metamorphic testing, we adopt two MRs, given a certain scenario during the daytime. The condition in MR1 is that the AV decision is unchanged regardless of day or night-time. For MR2, it requires the decision to be robust regardless of the time as well as the bad weather (intense rain and fog).

## IV. RESULT

### A. RQ1: Are our test cases effective in mimicking hazardous situations and can the system capture them?

During the preparation for Deliverable 1, we are yet to obtain any failure in six given scenarios in single tests. One may reckon that either (i) the system may not be able to record failures, (ii) these test cases are not diverse enough or (iii) a more refined setup for each test case is needed. Our result in this paper supports the latter options (i.e., (ii) and (iii)). In Deliverable 2, we have 4 classes of scenarios. We made the test cases more diverse by generating different scenarios associated with the parameterisation of pedestrians and NPCs involved. These values are parameterised to allow for greater testing coverage.

Our results show that there are 58 collisions found over the total number of 576 test cases, yielding a failure rate of 10.0% (Table I). This indicates that our test cases are effective in stimulating collisions in the system, and at the same time, the Apollo can capture it well. In addition, it is found that some classes of test cases are more effective than others. For example, in class A, it is shown that the failed cases are 31.0% (=25/81) as shown in Table I. In comparison, the class C, the Apollo drives the ego car so well that it does not cause any collision (Table I), even in case the headway car is approaching on the same side of the road but in a different lane (Figure 3).

TABLE I. SUMMARY OF ALL SCENARIOS

| Scenarios | | Numbers | | |
|---|---|---|---|---|
| Class | Code name | Parameters | Failed cases | Total cases |
| A | Close Quarters | 4 | 25 | 81 |
| B | Pedestrian at Intersection | 6 | 28 | 216 |
| C | Go Around, Oncoming | 3 | 0 | 36 |
| D | Camera Tricks | 5 | 5 | 243 |
| | Total | 18 | 58 | 576 |

### B. RQ2: What are critical factors that distinguish a scenario to be safe or unsafe?

*a) Scenario Class A (Pedestrian in Close Quarters):* In this class of scenario (Figure 5a), the ego vehicle is set to drive to the next intersection, just beyond the line of N cars (10, by default). The pedestrians spawn in the place of the ith (default, 5th) car position, with waypoints set to the position in the opposite line of cars, along with a trigger distance of initially 20 meters. These values are parameterised to allow for greater testing coverage. We found that the main factor that affects the safety is the distance where the pedestrian starts moving (pedTrigger). Most collisions (76.0%) occur when the pedestrian starts within 10 metres from the distance to the ego car (Table II). When the pedestrians start at 20 metres, the collision rate drops to 24.0% (Table II). No collision occurs when the trigger distance is at 30 metres. It is found that the place where the pedestrian starts crossing the road (pedLocation) also slightly affects the collision rate, with about half (48.0%) of the collisions occurring at the location of the third car (Table II). Given the fixed number of pedestrians, their speed (pedSpeed) and the number of cars on the road (numberOfCar) seem not to be the main factors affecting the system decision in this particular scenario. The relationships between parameters are represented in Figure 6a. The strongest relationship is the configuration between the nodes (pedSpeed=4, pedTrigger=10, numberOfCars=20, pedLocation=3). In other words, the most hazardous scenario in this class is associated with the situation in which the pedestrian moves at the speed 4 m/s, being triggered within 10 metres from the ego car at the location that is closed to the third NPC in the setting with a total number of 20 NPCs.

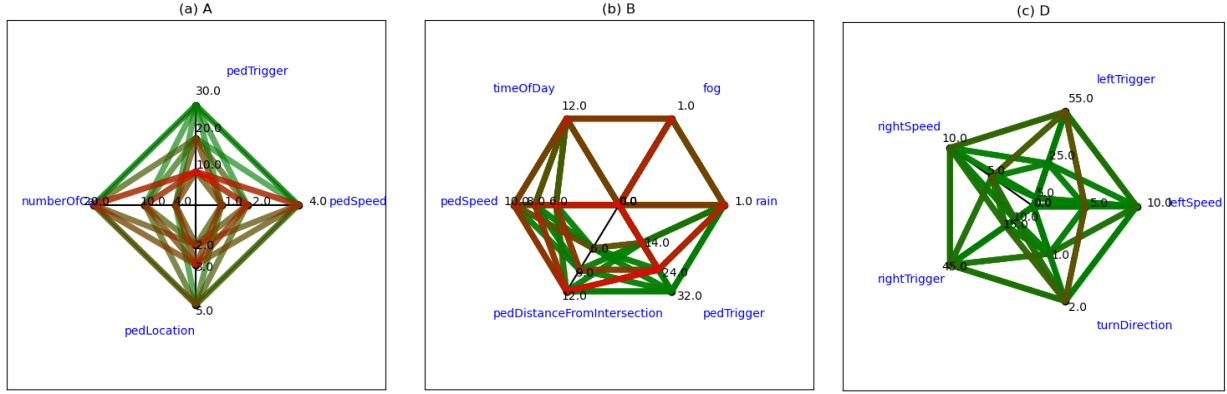

Fig. 6. Combined radar plots for representing relations of parameters: (a) A (Pedestrians in Close Quarters), (b) B (Pedestrians at Intersection), (c) D (Camera Tricks at Intersection). Green and red colors are to represent the safe and unsafe scenarios, respectively. Parameter value in each axis is in black color.

TABLE II. SUMMARY OF PARAMETERS OF FAILED SCENARIOS A

| Name of parameter | Value | Failure scenarios | |
|---|---|---|---|
| | | Total | Percentage |
| pedSpeed | 1 m/s | 7 | 28.0% |
| | 2 m/s | 10 | 40.0% |
| | 4 m/s | 8 | 32.0% |
| pedTrigger | 10 m | 19 | 76.0% |
| | 20 m | 6 | 24.0% |
| | 30 m | 8 | 0.0% |
| numberOfCar | 4 cars | 7 | 28.0% |
| | 10 cars | 8 | 32.0% |
| | 20 cars | 10 | 40.0% |
| pedLocation | 2-th car | 8 | 32.0% |
| | 3-th car | 12 | 48.0% |
| | 5-th car | 5 | 20.0% |

*b) Scenario Class B (Pedestrian at Intersection):* Different to the first class of scenarios, an intersection with a turn is involved and the ego expects to have complete right of way (Figure 5b). Again, we found that the distance (and thus the time) for which the pedestrian starts moving is the most critical factor that affects the collision between them and the ego car. The majority (85.7%) of incidents happen at a trigger distance of 24 metres (Table III). When the trigger distance (pedTrigger) is larger or smaller (i.e., the timing when the ego car encounters the pedestrian is later or earlier) there are much fewer collisions. The distance from the initial position of pedestrian to the crosswalk is also an important factor. The weather (rain and fog) and the time of the day (timeOfDay) do not have a clear impact on the collision rate (Table III) in this scenario. The most hazardous scenario in this class is associated with the configuration in which the pedestrian moves at the speed 10m/s (pedSpeed), being triggered at the distance 24 metres from the ego car (pedTrigger) and 12 metres far away from the intersection (pedDistanceFromIntersection) as indicated in Figure 6b.

TABLE III. SUMMARY OF PARAMETERS OF FAILED SCENARIOS B

| Name of parameter | Value | Failure scenarios | |
|---|---|---|---|
| | | Total | Percentage |
| rain | 0 (no fog) | 15 | 53.6% |
| | 1 (intense) | 13 | 46.4% |
| fog | 0 (no fog) | 14 | 50.0% |
| | 1 (intense) | 14 | 50.0% |
| timeOfDay | 12 (noon) | 13 | 46.4% |
| | 0 (midnight) | 15 | 53.6% |
| pedSpeed | 6 m/s | 8 | 28.6% |
| | 8 m/s | 8 | 28.6% |
| | 10 m/s | 12 | 42.8% |
| pedTrigger | 14 m | 4 | 14.3% |
| | 24 m | 24 | 85.7% |
| | 32 m | 0 | 0% |
| pedDistanceFromIntersection | 6 | 4 | 14.3% |
| | 9 | 7 | 25.0% |
| | 12 | 17 | 60.1% |

*c) Scenario Class C (Go Around, Oncoming):* There are several parameter combinations in this class; however, no collision occurs between cars (Table I). There is no pedestrian in this scenario. It is interesting to note that the ego vehicle does not even attempt to go around the blocked car into the other lane, suggesting that the ego vehicle does not want to encroach onto the other side of the road, despite the lane lines (Figure 5c). We are not familiar with road rules in the US and the line markings are not very clear in the map itself, but it would seem that this particular road should allow vehicles to drive on the other side for overtaking and passing.

*d) Scenario Class D (Camera Tricks at Intersection):* It was found from this scenario that the ego will not stop in time when set to drive straight through the intersection (Figure 5d), colliding with at least one NPC vehicle in some scenarios (Table IV). This is likely because the ego always has right of way, regardless of its destination direction, however, when it is driving straight, it does not need to slow down so cannot respond to the encroaching perpendicular vehicles. When turning, however, the ego will slow down before the turn, allowing it to detect and avoid the NPCs. Despite this, it can still be possible for the ego to have a collision, even when making a turn. Among the factors that determine the high collision ratios, the turn direction is the most important one. The majority (4 out of 5) of collisions are found when the ego car attempts to turn left (Table IV). The timing associated with the car on the left (leftTrigger) is also an important factor, as all the collisions occur when it is triggered at a distance of 55 metres (Table IV). The speed of the leftside car is also important. In contrast, the variations in speed and distance of rightside car (blue colour in Figure 5d) do not make any significant difference. The most hazardous scenario in this class is associated with the configuration in which the ego car turn left (turnDirection=2) in which the left car starts moving at the distance of 55 metres (leftTrigger) with the car on the right (rightSpeed) having the speed of 5 m/s (instead of zero as in Table IV) as indicated in Figure 6c.

TABLE IV. SUMMARY OF PARAMETERS OF FAILED SCENARIOS D

| Name of parameter | Value | Failure scenarios | |
|---|---|---|---|
| | | *Total* | *Percentage* |
| leftSpeed | 0 m/s | 0 | 0.0% |
| | 5 m/s | 4 | 80.0% |
| | 10 m/s | 1 | 20.0% |
| leftTrigger | 5 m | 0 | 0.0% |
| | 25 m | 0 | 0.0% |
| | 55 m | 5 | 100.0% |
| rightSpeed | 0 m/s | 3 | 60.0% |
| | 5 m/s | 2 | 40.0% |
| | 10 m/s | 0 | 0.0% |
| rightTrigger | 10 m | 1 | 20.0% |
| | 15 m | 2 | 40.0% |
| | 45 m | 2 | 40.0% |
| turnDirection | 0 (go straight) | 1 | 20.0% |
| | 1 (turn left) | 0 | 0.0% |
| | 2 (turn right) | 4 | 80.0% |

### C. *RQ3: Does the environmental condition have any impact on safety?*

In the class of scenario B, there is no clear impact of weather on the collision rates, most likely since the test cases are driven by multiple parameters simultaneously, in which some parameters are more important than others in the AV's decisions. To make it more objective on the impact of weather and light conditions, we applied the metamorphic testing for the class of scenario D. We selected this class because it is the case for which the collision rate is neither largest nor smallest. It is good to see if there any "improvement" in the result.

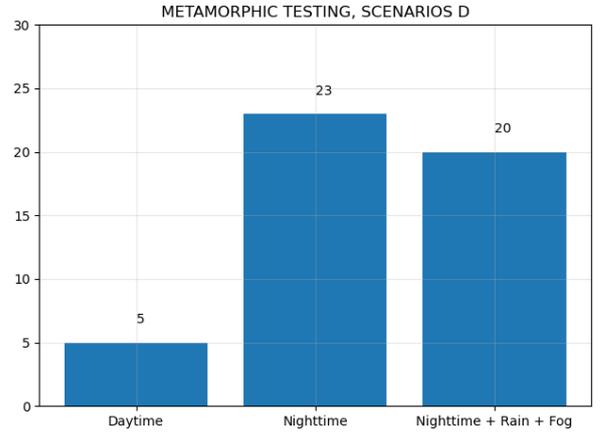

Fig. 7. Summary of failed outputs for the class of scenarios D. For the metamorphic testing, source input is implemented in day time and good weather (no rain, no fog). In the follow-up input of MR1, we adopt the change to be night-time. In the follow-up input of MR2, the environment is set with night-time in an intense rainy and foggy weather.

For this purpose, we carried out additional experiments for the class of scenario D that is listed in Table I. We adopt metamorphic testing, in which the source input is implemented in the daytime and good weather (no rain, no fog). In the MR1, we config the follow-up inputs to be the tests at night-time. In the MR2, the follow-up inputs are for the environment of night-time in intense rainy and foggy weather. Note that in scenario class D, NPC cars are set as black so that it may cause some difficulties for the system to detect them. In total, we have new 486 follow-up scenarios being generated from 243 source scenarios.

The system is expected to be robust against a wide range of weather and light conditions. However, MT helps us to determine that this is not the case with scenario D. There are only 5 failures over the total of 243 source test cases. However, in both MR1 and MR2, the number of failures increases by 4 times or higher. The result indicates that the environmental condition does have an impact on safety.

Interestingly, the failure rate of the worst condition (night-time in intense rainy and foggy weather) is 8.2% (=20/243) lower than the medium condition (night-time) as of 9.5% (=23/243). This shows that either the system is indeterministic with large uncertainty, or something related to the physics has not been explored, such as the black cars may be more visible under rains during nighttime other than the pitch-black sky.

### V. DISCUSSION

In this section, we discuss a general question on the use of a simulator-based system based on our results with the SVL/Apollo system, that is:

*RQ4: Is the use of simulator-based systems helpful in understanding and guaranteeing the safety of AVs in reality?*

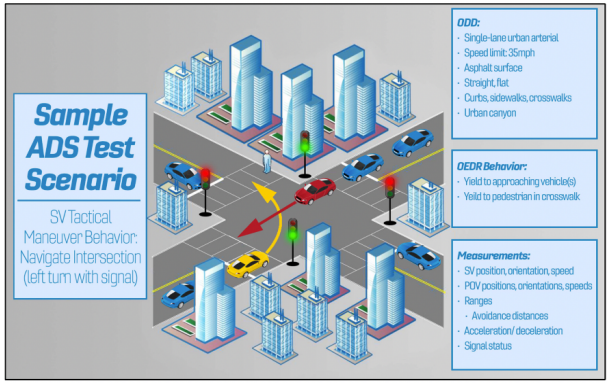

Fig. 8. A suggested sample test scenario for AV (after [9]).

The development of safety standards for autonomous vehicles is only in its infancy. At present, the closest standard is ISO 26262:2018 [6], which is applicable to generic computer-based system safety for conventional vehicles that requires a certain level of human intervention. It was pointed out that such a standard may not be applicable to autonomous vehicles, and a better approach is needed to fully verify and validate the AVs [7]. Several recent studies have proposed certain scenarios and test cases for testing AVs [8, 9]. However, it remains unknown how to implement the testing systematically, either with simulator-based testing or with road tests.

Our study demonstrates that simulator-based systems are helpful in understanding and guaranteeing the safety of AVs. First, all our test scenarios are derived from naturalistic driving situations (Figure 1—4). At the same time, our test scenarios B (Figure 5b) and D (Figure 5d) are similar to the setup of sample test scenarios (Figure 8) suggested by previous studies [9]. In other words, they can be applied to guarantee the safety level once the AV passes the test cases. Secondly, the effectiveness of Apollo in detecting unsafe test cases in the SVL simulator is close to human intuition. For example, as discussed (in Section IV.B), it is found that the main factor that affects safety is the distance (and thus time) where the pedestrian starts moving across the road. The closer he/she starts to move when the ego car approaches, the higher risk the pedestrian faces. On the other hand, the weather and light condition do have an impact on safety (in Section IV.C). During night-time with or without bad weather, the system makes more bad decisions compared to the daytime. Qualitatively, it is similar to human decisions because intuitively we will make more mistakes in driving at night rather than during daytime. Note that: human makes only about 2 times more mistakes during driving at night than during daytime [10]. In addition to the difference in the capability of drivers, it is also to the fact that the comparison is with differences in scenarios and setups.

It is reported in our study that collisions occur in 10.0% of all tested scenarios. Considering that there are many millions of people involved in the traffic every day, this ratio is unacceptably large. As indicated earlier, deep learning systems that drive AVs with simulator-generated scenes are as accurate as the testing with real-life datasets [3]. If the same system is to be used in the real world, more comprehensive and systematic tests are of great necessity to ensure that they are reliable, robust, and safe at a certain level before being allowed on our roads.

## VI. CONCLUDING REMARKS

Safety is one of the main challenges that determine the acceptance of AVs to be on the road. Simulators enable us to test the performance of AVs conveniently, efficiently, and safely. In this paper, we present an investigation using a simulator to study the behaviours of an AV system. We have adopted the SVL simulator in combination with the Apollo platform to carry out the simulations. Our scenarios are inspired by real-life driving situations. We have used EP and MT testing methods to evaluate a total number of 1062 test cases.

Our main findings are as follows. (1) Our proposed scenarios are effective in mimicking real-life situations; and at the same time, the system can capture collisions well; (2) Several critical factors distinguish a scenario to be safe or unsafe; (3) The environmental conditions do have an impact on the safety of AVs; and (4) The use of simulator-based system helps understand the safety in reality. Our study provides new empirical evidence to support the argument that AV simulator systems may well represent and capture real-life scenarios. If the system performs similarly in reality, our study implies that state-of-the-art AVs should be tested thoroughly before deployment to reduce safety-critical problems.

## ACKNOWLEDGEMENT

We appreciate the encouragement and support of Prof. Tsong Yueh Chen, Dr. Huai Liu (Swinburne University of Technology), and Prof. Hai L. Vu (Monash University). We also would like to thank the Organising Committee of IEEE Autonomous Driving AI Test Challenge for helpful training and support, as well as two reviewers for useful comments.